# Using Data Analytics to Derive Business Intelligence: A Case Study

**Ugochukwu Orji[1]; Ezugwu Obianuju[1]; Modesta Ezema[1]; Chikodili Ugwuishiwu[1]; Elochukwu Ukwandu[2,*]; Uchechukwu Agomuo[1]**

[1]Department of Computer Science, Faculty of Physical Science, University of Nigeria, Nsukka, Enugu State, Nigeria

[2]Department of Applied Computing and Engineering, Cardiff School of Technologies, Cardiff Metropolitan University, Wales, United Kingdom.

Email addresses: {ugochukwu.orji.pg00609; assumpta.ezugwu; modesta.ezema; chikodili.ugwuishiwu; uchechukwu.agomuo.pg00090}@unn.edu.ng, eaukwandu@cardiffmet.ac.uk

[*]Correspondent Author: Elochukwu Ukwandu {eaukwandu@cardiffmet.ac.uk, ORCID: 0000-0003-1350-4438}

**Abstract**

The data revolution experienced in recent times has thrown up new challenges and opportunities for businesses of all sizes in diverse industries. Big data analytics is already at the forefront of innovations to help make meaningful business decisions from the abundance of raw data available today. Business intelligence and analytics (BIA) has become a huge trend in today's IT world as companies of all sizes are looking to improve their business processes and scale up using data-driven solutions. This paper aims to demonstrate the data analytical process of deriving business intelligence via the historical data of a fictional bike-share company seeking to find innovative ways to convert their casual riders to annual paying registered members. The dataset used is freely available as "Chicago Divvy Bicycle Sharing Data" on Kaggle. The authors used the R-Tidyverse library in RStudio to analyze the data and followed the six data analysis steps of; ask, prepare, process, analyze, share, and act to recommend some actionable approaches the company could adopt to convert casual riders to paying annual members. The findings from this research serve as a valuable case example, of a real-world deployment of BIA technologies in the industry, and a demonstration of the data analysis cycle for data practitioners, researchers, and other potential users.

**Keywords:** Data Analytics, Data Analysis Cycle, Business Intelligence, Big Data Analytics

## 1.0 Introduction

The continuous advancements in technological innovations are facilitating the enormous growth of heterogeneous data from multiple sources; thus, creating new challenges and even bigger opportunities for businesses and academics. These data being generated are both structured and unstructured, complex and simple, creating the boom of big data. Furthermore, nowadays, there are smart gadgets, the Internet of Things (IoT), and user-generated content (UGC) from social media, all contributing to the large data pool.

For academia, the dynamic nature of the data being generated is creating unprecedented research opportunities in several fields, including; social science, economics, finance, biology & genetics, etc. [1].

Data analytics has become a key enabler for business transformation for many businesses, no matter their size, and has already changed the way businesses operate through better customer service, payments, business models, and new ways to engage online [2]. In order to generate actionable insights, large volumes of structured/unstructured data can be analyzed from multiple sources using data analytics for business intelligence [3]. This ability to convert vast amounts of opaque data into refined and action-driven information in real-time offers a significant competitive advantage to these businesses and BIA is the leading technology driving this innovation.

Orji et al. [4] detail how organizations of all sizes are currently utilizing business intelligence tools and techniques to explore result-oriented ways to provide business value and improve decision-making for their businesses. According to Hočevar & Jaklič [5], for organizations to thrive and stay competitive, effective and timely business information is key, especially in today's rapidly changing business environment. Managers who are serious about maintaining their business competitiveness in this age of Digital Business Transformation (DBT) must not rely solely on their intuition or other unconventional business approaches. An organization's decision-making process no matter the size must be data-driven. This is especially important because a vital component of strategic planning and management for any organization is data. With data, they can analyze their strengths and weaknesses, those of their competitors, and then anticipate market developments and predict their competitive environment [3] [6].

A major industry where data analytics has widely and successfully influenced their mode of operation is e-commerce and e-services [7]. A TDWI survey of 2009 showed that 38% of the surveyed businesses and organizations are already practicing an advanced form of data analytics, and another 85% indicated their intentions of deploying it in the future [8]. According to a 2020 Statista forecast [9], the BIA software applications market size will see an enormous worldwide increase over a six years stretch. It is estimated to grow from $14.9 billion in 2019 to $17.6 billion in 2024. A different report by NewVantage Partners [10] suggests that 91.6% of Fortune 1000 companies are investing in Big Data Analytics (BDA) with 55% of firms already investing more than $55 million. This shows great opportunity in BIA.

## 1.1 Research Objectives and Questions

The objective of this research is to derive business intelligence using data from a 'real-life' situation. Adopting the six data analysis steps, we examined data from a fictitious bike-share company (CYCLISTIC BIKE SHARE) over a period of one year; October 2020 to September 2021. The aim of the project is to design marketing strategies aimed at converting the company's casual riders into annual paying members using the historical dataset.

To achieve this objective, the following are some key issues to resolve from the data;
  i. Understand the riding pattern of annual members and casual riders
  ii. Find opportunities to convert casual riders to paying annual member
  iii. Find possible digital marketing strategies to convert casual riders into members

## 2.0 Review of Related Works

This study investigated some available literature where data analytics was utilized in deriving business intelligence for various businesses in different industries.

The opportunities and possibilities opened as a result of data analytics are enormous. Malhotra & Rishi [11] explored these possibilities for e-commerce using an analysis of customer preferences and browser behaviors. They found a way to ensure seamless online purchase decisions with the aid of personalized page ranking order of web links from customer queries. Their research aimed to find solutions to the limitations of available search and page ranking systems of e-commerce platforms.

Increasingly, organizations are tapping into the potent power of social media to grow their business, gauge user sentiments on their products and services and stay ahead of trends. This is especially vital to small and medium enterprises (SMEs) as they seek to break even bigger competitors. In [12] Orji et al. developed a framework where SMEs can see user sentiments on their products and services based on data generated from Twitter. This is regarded as "crowd wisdom" and helps businesses streamline their business processes accordingly to meet the needs of their customers.

Information has proven to be an essential resource for better decision-making and implementing robust business strategies. Caseiro & Coelho [13] while investigating the direct and indirect effects of Business Intelligence on organizations' performance, surveyed 228 startups across European countries. Their result indicated the apparent influence Business Intelligence capacities have on network learning, innovativeness, and performance. The authors highlighted the need for startups to pay attention to business intelligence capacities, given their impact on overall performance.

Hopkins & Hawking [14] used an intrinsic case study approach to examine the role and impact of BDA and IoT, in supporting the strategy of a large logistics firm to improve driver safety, reduce operating costs, and reduce the environmental impact of its vehicles. Based on the results of the BDA, truck routing will be improved, along with recommendations for optimal fuel purchases and locations, and a forecast for proactive and predictive maintenance schedules.

Furthermore, below are practical use cases where Data Analytics have been effectively utilized to derive Business Intelligence in various industries.

**Table** 1: different practical use cases of data analytics to improve business intelligence

| Businesses | Impact of data analytics on the business intelligence process | References |
|---|---|---|
| Netflix | With its 151 million subscribers, Netflix implements data analytics models to determine customer behavior and buying patterns and then recommends movies and TV shows based on that information. | [15] |
| Express scripts | Analyze patient data and alert healthcare workers to serious side effects before prescribing medications | [16] |
| McDonald's Corporation | Based on McDonald's Corporation sales data, the drive-through experience, kitchen operations, supply chain, menu suggestions, personalized menus, and deals are optimized. | [17] |
| CitiBank | The online banking provider help minimize financial risk by analyzing big data and pinpointing fraudulent behaviors using real-time machine learning and predictive modeling. Users can be notified of suspicious transactions, for example, incorrect or unusual charges, promptly by CitiBank. Besides being helpful for consumers, this service is also useful for payment providers and retailers in monitoring all financial activity and identifying threats to their businesses. | [18] |
| Advanced Radiology Services | This private radiology practice built a data warehouse to improve its daily practice management. With over 100 radiologists in a wide area, the organization had limited data on which to base decisions before the DW was implemented. By investing in IT infrastructure they have since seen an increase | [19] |

| | of 10.4% in productivity in the first two years and experienced growth in existing sites and acquired new contracts. | |
|---|---|---|
| Various Auditors | Auditors are using data analytics to perform audit procedures including:<br>• The metadata attached to transactions is used to identify combinations of users involved in processing transactions<br>• Analyzing revenue trends by product and region<br>• Matching purchase orders with invoices and payments<br>• NRV testing - comparing the price at which an inventory item was purchased and sold last time | [20] |

## 3.0 Research Methodology

This study followed the six data analysis cycle as shown below:
Ask→ Prepare → Process → Analyze → Share → Act
This section will deal with the first 4 stages of the data analysis cycle and the rest in the following sections.
a. Ask: At this stage of the data analysis cycle, the task is to understand the stakeholder expectations and plan for how to go about the remaining stages of the process. The key tasks here are:
- Identify the business task – this is the aim of the project which is to analyze the Cyclistic historical bike trip data to identify trends that will help the marketing team convert casual riders into annual paying members.
- Consider key stakeholders – in this case study, the primary stakeholder would be the director of marketing, while the secondary stakeholders include; the marketing analytics team and the executive team of the Cyclistic company.

Figure 1 shows the steps taken to achieve our data analysis objectives in this research

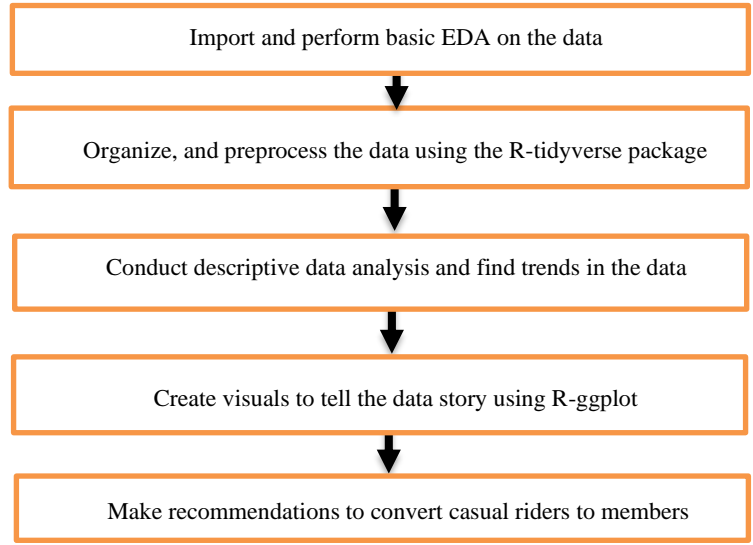

Figure 1: Research flow

b. Prepare: At this stage of the process, the task is to prepare the data for further analysis and, in doing so, must consider the following:
- Choose data sources – the dataset consists of historical trip data of the Cyclistic bike-share company recorded between October 2020 and September 2021. The data has been made available by Motivate Inc. and is publicly available as "Chicago Divvy Bicycle Sharing Data" on Kaggle [21].
- To begin, the analyst loads the data into the RStudio environment, then performs some basic Exploratory Data Analysis (EDA) to better understand the data. We inspect the data to make sure that the data contains the right information needed for the task at hand. Here we check the column names, data types, and overall consistency of the data.

**Table** 2: Brief description of the dataset

| S/N | Attribute | Description | Data type |
|---|---|---|---|
| 1 | ride_id | User Ride ID | String |
| 2 | rideable_type | Type of Bike (classic_bike, electric_bike & docked_bike) | String |
| 3 | started_at | Trip Start Date & Time | Date |
| 4 | ended_at | Trip End Date & Time | Date |
| 5 | start_station_name | Trip Start Station Name | String |
| 6 | start_station_id | Trip Start Station ID | String |

| 7 | end_station_name | Trip End Station Name | String |
|---|---|---|---|
| 8 | end_station_id | Trip End Station ID | String |
| 9 | start_lat | Trip Starting Latitude | Numeric |
| 10 | start_lng | Trip Starting Longitude | Numeric |
| 11 | end_lat | Trip Ending Latitude | Numeric |
| 12 | end_lng | Trip Ending Longitude | Numeric |
| 13 | member_casual | Registration Type (member or casual) | String |

c. Process: At this stage of the data analysis cycle, the data is cleansed and pre-processed for analysis. The following data cleansing and preprocessing procedures were taken:
- Remove duplicates: we first removed duplicates from the dataset
- Remove rows with missing values: next we handle missing values
- Parse datetime columns: i.e. convert the string representation of the date and time value to its DateTime equivalent so it stacks correctly.
- Next we manipulate the data further by adding new columns that will help improve calculations e.g. new columns to list the weekday, month, and year of each ride which will be useful to determine users' riding patterns. We also added new columns to calculate each ride length in hours which will be useful for intra-day analysis.
- Finally we removed the geographic coordinates date (Longitude & Latitude) because it was not needed for our analysis.

d. Analysis: At this stage of the data analysis cycle, the data analyst performs a descriptive analysis of the cleaned data. The key tasks at this stage include:
- Performing summary statistical calculations: summary statistics is done to provide a quick and simple description of the data.
- Aggregating the data so that it's useful and accessible: here, the data is sorted, normalized, and grouped accordingly to get insights.
- Identifying trends and relationships: this is the main aim of the data analysis process, to see the trends, patterns, and relationships in the data.

### 4.0 Result and Discussion

e. Share: After analyzing the data and identifying the trends, the next stage of the data analysis cycle is to share the findings and tell the data story to stakeholders who have been looking forward to it. For this case study, the data analyst used the ggplot library inherent in the R-tidyverse package to create visualizations and help the stakeholders better understand the data. Below are some snippets of the visuals from the data analysis.

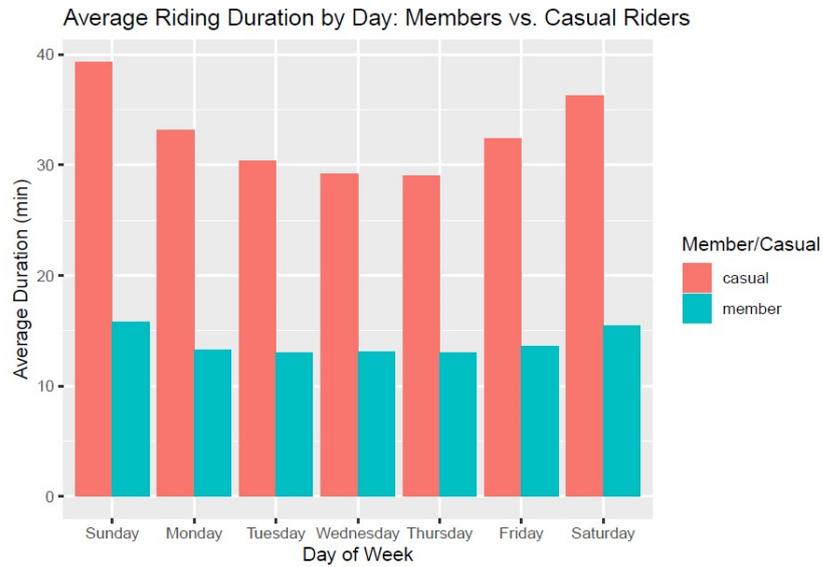

Figure 2: Average riding duration (in minutes) per day for both categories of riders

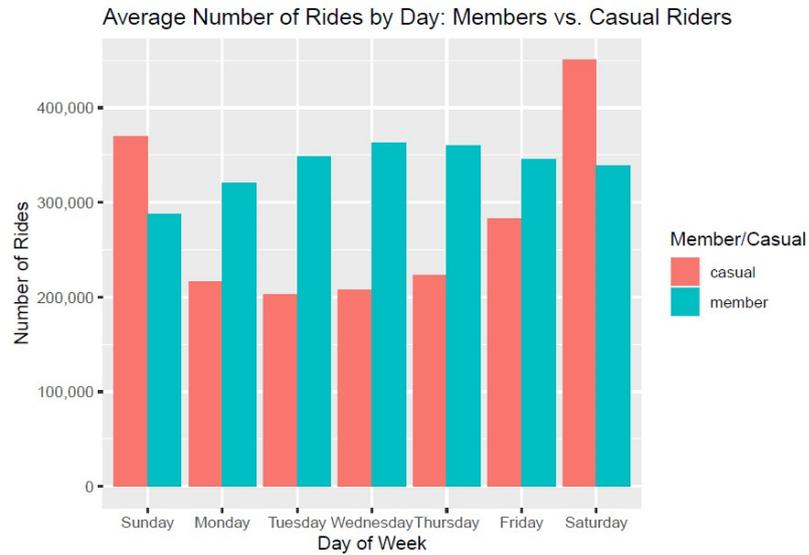

Figure 3: Average number of rides per day for both categories of riders

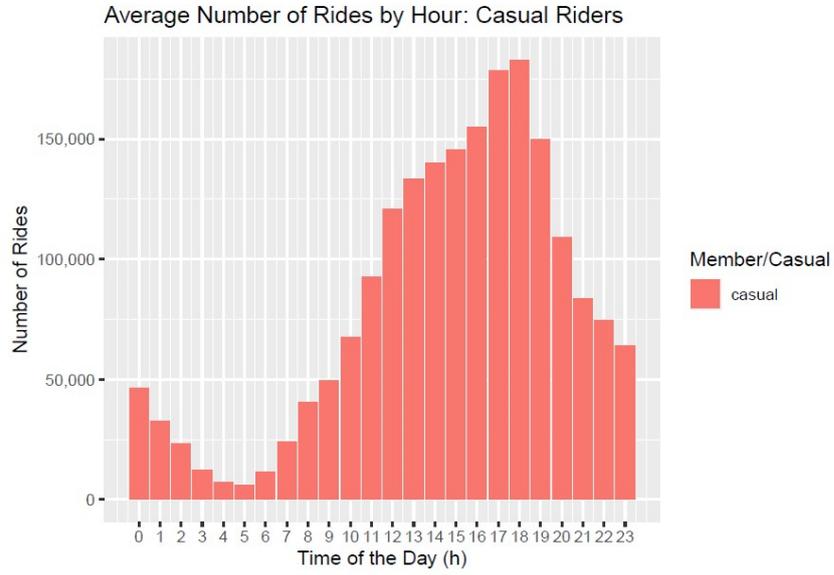

Figure 4: Average number of rides per hour for casual riders

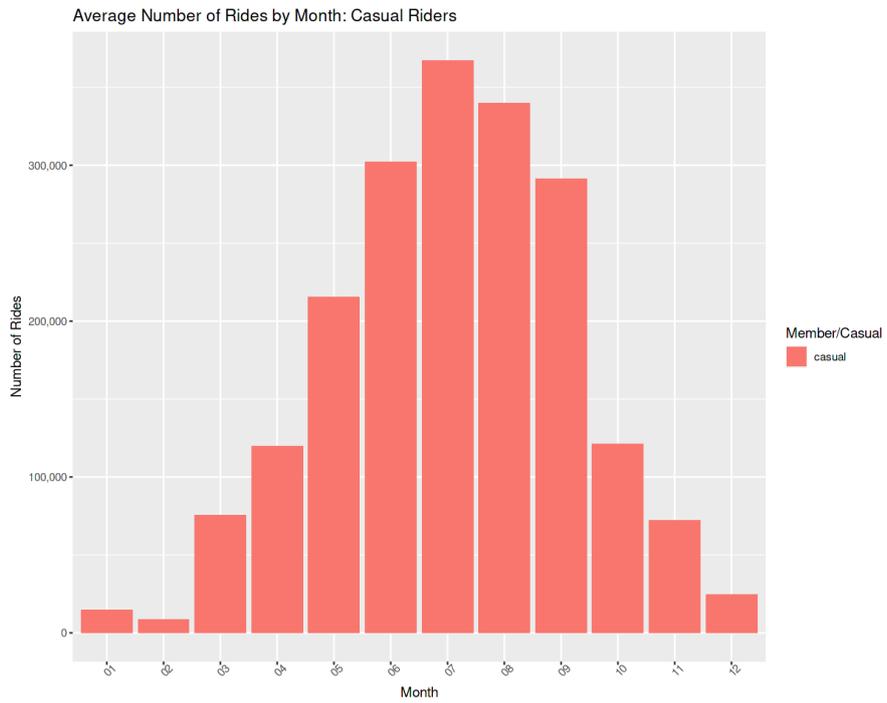

Figure 5: Average number of rides per month for casual riders

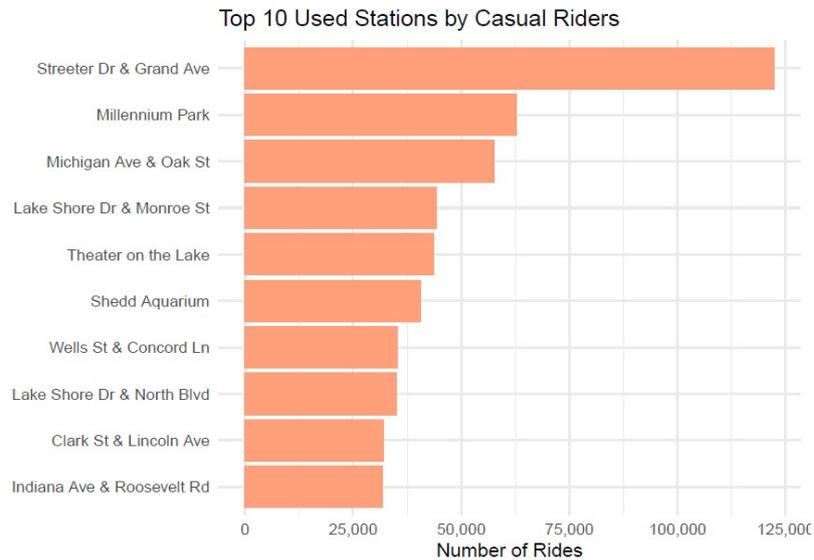

Figure 6: Top 10 most used stations by casual riders

Key takeaways:
- The average ride duration is higher for casual riders for all indices measured.
- Casual riders ride more during the weekends while annual paying members preferred the mid-week days.
- Unsurprisingly, the summer months of June to September were the peak riding months for casual riders.
- Streeter Dr & Grand Ave. is the most used bike station for casual riders with over 100,000 rides almost twice as any other bike station.

f. Act: Finally, at the Act phase of the data analysis cycle, the stakeholders are to be given actionable recommendations based on the findings from the data analysis. From our findings, the following recommendations were made to the stakeholders in their bid to convert casual riders to annual paying members:
  1. Give incentives to members and offer rewards for achieving set riding milestones to attract casual riders since they already have huge riding numbers.
  2. Offer occasional membership discounts to newly registered members in summer, holidays, and weekends since most casual riders prefer to ride during those periods.
  3. Partner with local businesses within the top 10 most used bike stations for casual riders especially Streeter Dr & Grand Ave. We recommend advertising with local businesses within the Streeter Dr & Grand Ave. bike station targeting local riders and frequent visitors (commuters).

## 5.0 Conclusion

In this paper, the researchers have used the Chicago Divvy Bicycle Sharing dataset to demonstrate the process of deriving business intelligence via data analytics tools and techniques using the case study of the fictional Cyclistic bike-share company. This paper also demonstrates why data analytics and business intelligence have emerged as the new frontier of innovation for businesses and startups looking to be competitive in today's markets. We believe our findings make a significant contribution, to an emerging research area that is currently lacking in academic research. We hope that this paper bridges the gap between industry practice and academic theory.

Data-driven decision-making is essential for businesses of all sizes, especially SMEs, in order to remain competitive. Powerful data analytics tools such as Python, R, Tableau, and Power BI make it easy to manipulate and visualize data. However, it is important to recognize that successfully leveraging data analytics requires more than just technology investments and experimentation with new techniques. Other key components include having a deep technical and managerial understanding of big data analytics capabilities, cultivating an environment of organizational learning, as well as integrating big data decision-making into a firm's operations. It is ultimately the combined effect of these resources that will help organizations develop their big data analytics capability and gain value.

## 5.1 Limitations and Future Works

Further analysis could be done to improve the findings of this study; for example, sourcing additional data like climate data and customer demographics data could help better understand the customer persona and provide more insights. Further analysis might also be needed to understand the motivations behind the increase in the number of causal riders on weekends, and whether medical or health-driven issues may have influenced this choice.

Future work should consider predictive analytics of the data to find patterns, identify risks, and opportunities.

## Additional Information

The datasets analyzed and complete documentation of the data analysis and programming process are available at: https://www.kaggle.com/orjiugochukwu/cyclistic-data-analysis

**References**

[1] Huang, S. C., McIntosh, S., Sobolevsky, S., & Hung, P. C. (2017). Big data analytics and business intelligence in industry. Information Systems Frontiers, 19(6), 1229-1232. https://doi.org/10.1007/s10796-017-9804-9
[2] Akter, S., Michael, K., Uddin, M. R., McCarthy, G., & Rahman, M. (2022). Transforming business using digital innovations: The application of AI, blockchain, cloud, and data analytics. Annals of Operations Research, 1-33. https://doi.org/10.1007/s10479-020-03620-w
[3] Mikalef, P., Boura, M., Lekakos, G., & Krogstie, J. (2019). Big data analytics and firm performance: Findings from a mixed-method approach. Journal of Business Research, 98, 261-276. https://doi.org/10.1016/j.jbusres.2019.01.044
[4] Orji, U. E., Ugwuishiwu, C. H., Nguemaleu, J. C., & Ugwuanyi, P. N. (2022, April). Machine Learning Models for Predicting Bank Loan Eligibility. In 2022 IEEE Nigeria 4th International Conference on Disruptive Technologies for Sustainable Development (NIGERCON) (pp. 1-5). IEEE. https://doi.org/10.1109/NIGERCON54645.2022.9803172
[5] Hočevar, B., & Jaklič, J. (2010). Assessing benefits of business intelligence systems–a case study. Management: journal of contemporary management issues, 15(1), 87-119.
[6] George, B., Walker, R. M., & Monster, J. (2019). Does strategic planning improve organizational performance? A meta-analysis. Public Administration Review, 79(6), 810-819. https://doi.org/10.1111/puar.13104
[7] Sun, Z., Zou, H., & Strang, K. (2015). Big data analytics as a service for business intelligence. In Conference on e-Business, e-Services and e-Society (pp. 200-211). Springer, Cham. https://doi.org/10.1007/978-3-319-25013-7_16
[8] Ram, J., Zhang, C., & Koronios, A. (2016). The implications of big data analytics on business intelligence: A qualitative study in China. Procedia Computer Science, 87, 221-226. https://doi.org/10.1016/j.procs.2016.05.152
[9] Statista. (2020, December 22). Global BI & analytics software market size 2019–2024. Retrieved Mar. 03, 2023 from https://www.statista.com/statistics/590054/worldwide-business-analytics-software-vendor-market/
[10] NewVantage Partners. (2019). Big data and AI Executive Survey 2019. Retrieved Mar. 03, 2023 from http://newvantage.com/wp-content/uploads/2018/12/Big-Data-Executive-Survey-2019-Findings-122718.pdf.
[11] Malhotra, D., & Rishi, O. (2021). An intelligent approach to design of E-Commerce metasearch and ranking system using next-generation big data analytics. Journal of King Saud University-Computer and Information Sciences, 33(2), 183-194. https://doi.org/10.1016/j.jksuci.2018.02.015
[12] Orji, U. E., Ezema, M. E., Ujah, J., Bande, P. S., & Agbo, J. C. (2022, April). Using Twitter Sentiment Analysis for Sustainable Improvement of Business Intelligence in Nigerian Small and Medium-Scale Enterprises. In 2022 IEEE

Nigeria 4th International Conference on Disruptive Technologies for Sustainable Development (NIGERCON) (pp. 1-5). IEEE. https://doi.org/10.1109/NIGERCON54645.2022.9803087

[13] Caseiro, N., & Coelho, A. (2019). The influence of Business Intelligence capacity, network learning, and innovativeness on startups performance. Journal of Innovation & Knowledge, 4(3), 139-145. https://doi.org/10.1016/j.jik.2018.03.009

[14] Hopkins, J., & Hawking, P. (2018). Big Data Analytics and IoT in logistics: a case study. The International Journal of Logistics Management. https://doi.org/10.1108/IJLM-05-2017-0109

[15] Dixon, M. (2019). How Netflix used big data and analytics to generate billions. Retrieved February 11, 2023 from https://seleritysas.com/blog/2019/04/05/how-netflix-used-big-data-and-analytics-to-generate-billions/

[16] Beall, A. (2020). Big data in health care: How three organizations are using big data to improve patient care and more? Retrieved February 11, 2023 from https://www.sas.com/en_gb/insights/articles/big-data/bigdata-in-healthcare.html.

[17] Elmes, S. (2019). Delicious Data: How big data is disrupting the business of food. Retrieved Mar. 03, 2023 from https://adimo.co/news/delicious-data-how-big-data-is-disrupting-the-business-of-food.

[18] Aleksandrova, M. (2019). Big data in the banking industry: The main challenges and use cases. Retrieved Mar. 02, 2023 from https://easternpeak.com/blog/big-data-in-the-banking-industry-the-main-challengesand-use-cases/

[19] Sigler, R., Morrison, J., & Moriarity, A. K. (2020). The importance of data analytics and business intelligence for radiologists. Journal of the American College of Radiology, 17(4), 511-514. https://doi.org/10.1016/j.jacr.2019.12.022

[20] Data analytics and the auditor | ACCA Global. Retrieved Mar. 03, 2023, from https://www.accaglobal.com/gb/en/student/exam-support-resources/professional-exams-study-resources/p7/technical-articles/data-analytics.html

[21] Chicago Divvy Bicycle Sharing Dataset. Retrieved from https://www.kaggle.com/datasets/orjiugochukwu/cyclistic-dataset